\documentclass[runningheads]{llncs}
\usepackage[T1]{fontenc}
\usepackage{graphicx}
\usepackage{xcolor}

\usepackage{multirow}
\usepackage{amsmath,amssymb,amsfonts}

\usepackage{array}

\newcolumntype{M}[1]{>{\centering\arraybackslash}p{#1}}

\begin{document}
\title{IHC Matters: Incorporating IHC analysis to H\&E Whole Slide Image Analysis for Improved Cancer Grading via Two-stage Multimodal Bilinear Pooling Fusion }

\author{Jun Wang\inst{1}\and
Yu Mao\inst{1}\and
Yufei Cui\inst{2}\and
Nan Guan\inst{1}\and
Chun Jason Xue\inst{3}
}
\authorrunning{F. Author et al.}
\institute{City University of Hong Kong \\
\email{jwang699-c}@my.cityu.edu.hk \and
McGill University\and
Mohamed bin Zayed University of Artificial Intelligence }

%\author{Jun Wang, Yu Mao, Yufei Cui, Nan Guan, Chun Jason Xue}
%\email{jwang699-c, yumao7-c}@my.cityu.edu.hk,yufei.cui@mail.mcgill.ca,nanguan@cityu.edu.hkjason.xue@mbzuai.ac.ae

\maketitle              % typeset the header of the contribution
\begin{abstract}

Immunohistochemistry (IHC) plays a crucial role in pathology as it detects the over-expression of protein in tissue samples. 
However, there are still fewer machine learning model studies on IHC's impact on accurate cancer grading. We discovered that IHC and H\&E possess distinct advantages and disadvantages while possessing certain complementary qualities. Building on this observation, we developed a two-stage multi-modal bilinear model with a feature pooling module. This model aims to maximize the potential of both IHC and HE's feature representation, resulting in improved performance compared to their individual use. Our experiments demonstrate that incorporating IHC data into machine learning models, alongside H\&E stained images, leads to superior predictive results for cancer grading. The proposed framework achieves an impressive ACC higher of 0.953 on the public dataset BCI.

\end{abstract}

%\subsubsection{Sample Heading (Third Level)} 
%\paragraph{Sample Heading (Fourth Level)}

\section{Introduction}

%Hematoxylin and eosin (H\&E) staining, a widely used and cost-effective technique in clinical settings, is pivotal for examining tissue samples. The application of deep learning methodologies in cancer classification tasks, particularly in the analysis of whole slide images (WSIs) stained with H\&E, holds significant promise. However, the utilization of H\&E staining comes with inherent limitations. The characteristic features of H\&E-stained WSIs, characterized by low contrast and varying shades of dark blue and pink, pose challenges for efficient digital image analysis and model representation learning.

Hematoxylin and eosin (H\&E) is widely used in clinical practice due to its accessibility and cost-effectiveness. Recent advancements in deep learning techniques have shown promise in cancer classification, particularly in analyzing H\&E whole slide images. However, analyzing these images digitally poses challenges due to their inherent characteristics. H\&E-stained whole-slide images lack molecular data associated with cells and are characterized by low contrast and varying shades of dark blue and pink. These factors make it difficult for the model to efficiently learn representations during digital image analysis\cite{dimitriou2019deep}.

This limitation can be overcome by utilizing the immunohistochemistry (IHC) staining technique, which can visualize protein localization in tissue samples, aiding tumor classification and detecting small clusters of tumor cells. IHC also provides predictive and prognostic information for breast cancer. However, IHC also has some limitations. 
The process of IHC is expensive, labor-intensive, and time-consuming compared with H\&E approaches. IHC labeling requires quantifying specific positive cells or structures, not the entire image~\cite{cizkova2021comparative}. Manual determination of regions of interest (ROIs) is necessary for accurate results. However, manual annotation is subjective, time-consuming, and costly due to challenges in distinguishing different cell types and setting thresholds for classification.
% Labeling of IHC requires quantifying a certain type of positive cells or structures, not the whole image. Such an example is Hofbauer placental cells, which show positivity of some antigens together with trophoblast, but only Hofbauer cells represent the regions of interest (ROIs). Thus, the precise manual determination of ROIs is necessary. A manual observer has the advantage of identifying technical staining errors or artifacts, but it is both time-consuming and costly. Additionally, manual annotation is error-prone and poorly reproducible, as histological samples consist of a mixture of different cell types that can be challenging to distinguish even by a trained eye, and setting a manual threshold of what is regarded as negative/positive is tedious and highly difficult. %Moreover, IHC WSI relies more on staining protocol.

\begin{figure}[t]
\includegraphics[width=\textwidth]{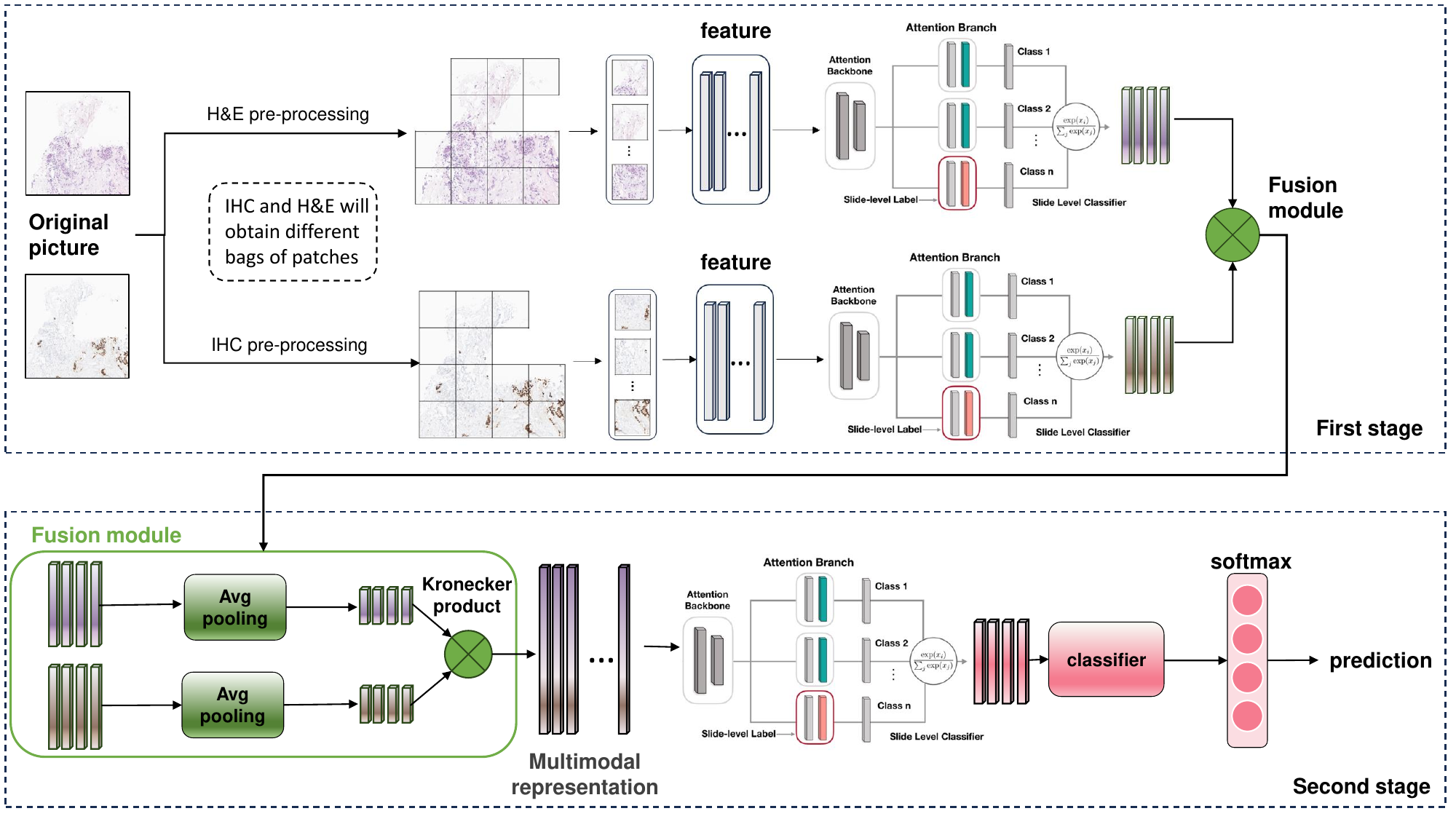}
\caption{Our framework: two-stage multimodal bilinear pooling Framework} \label{fig1}
\end{figure}

We propose a two-stage multimodal bilinear pooling fusion framework to fully utilize the advantages of both H\&E and IHC whole slide images. In our research, we employ Bilinear Average Pooling and Kronecker products to combine bag-level features from the H\&E-based and the IHC-based pipeline. The resulting merged feature is then utilized to train the machine learning model, aiming to enhance the accuracy of cancer grading predictions. The framework is illustrated in Fig.~\ref{fig1}. Our method achieves an accuracy score of 0.953 in breast cancer grading tasks, surpassing the performance of using either method independently. We demonstrate that combining histopathological images enhances performance and allows the model to effectively capture relevant information.

%Therefore, we combine H\&E with IHC whole slide images together as input by our proposed two-stage multimodal bilinear pooling fusion framework
%to train the machine learning model to achieve better prediction in cancer grading tasks. In our paper, we apply Bilinear Average Pooling and Kronecker products to fuse bag-level features from H\&E-based pipeline and IHC-based pipeline. Our Framework is shown in Fig.~\ref{fig1}). Our method ultimately achieved an accuracy (ACC) score of 0.953 in breast cancer grading tasks, surpassing the performance of using either method independently. Our analysis comparing the use of H\&E WSIs and IHC WSIs for classification has demonstrated that combining histopathological images enhances performance and enables the model to effectively capture relevant information.

\section{Related work}

\subsubsection{IHC}

Tranditional image processing method and IHC-based machine learning focus on color vector extraction~\cite{van2017image,shi2016automated,geread2019ihc} and quantification of intensity features~\cite{mukundan2019analysis}. %However, these tranditional methods can not achieve high ACC than machine learning model, due to the inherent characteristics of IHC-stained whole-slide images. Cells with deep color is not always strongly related to cancer. Some normal cell is also be stained. In placenta tissue, where trophoblast and Hofbauer cells both are IHC positive but only Hofbauer cells represent the regions of interest (ROIs)~\cite{cizkova2021comparative}.
Some previous work~\cite{bai2023deep,weitz2023multi,bci} focus more on Image translation and whole-slide-image registration algorithms. They believe that it is necessary to make pixel-to-pixel matches between H\&E and IHC images for further training.
%ACROBAT 2023~\cite{weitz2023multi} is hold for advancing the development of whole-slide-image registration algorithms that can align the H\&E WSIs with IHC WSIs from the same tumour sample. Liu et al.(2022)~\cite{bci} hold BCI challenge and propose a pyramid pix2pix method to  generate IHC data directly with the paired H\&E stained images.
Also, many other work~\cite{naik2020deep,anand2021weakly,he-her2net} think IHC dataset is hard to get so it is necessary to skip IHC images and predict cancer and molecular marker only based on H\&E whole slide images. 

%\subsubsection{Attention-based multiple instacen learning}Ilse et al.(2018)~\cite{ilse2018attention} firstly propose an attention-based multiple instance learning approach. Based on their work,  Lu et al.(2021)~\cite{lu2021data} first propose a novel interpretable model, which is called Clustering-constrained Attention Multiple instance learning (CLAM), to solve multi-class cancer classification problems. %The attention network of CLAM returns scores for each patch and ranks those patches by their attention scores. The model selects the top-k most important patches to learn the representative features.

\subsubsection{Multimodal}

Bilinear pooling model have been widely used in other fields for multimodal tasks.
Kim et al.(2016)~\cite{kim2016hadamard} introduce Hadamard product into the bilinear pooling model to construct the multimodal features for visual question-answering tasks. Lin et al.(2017)~\cite{lin2015bilinear} design a bilinear layer and a pooling layer in CNN to generate features from two CNN and subsequently integrate these features via the outer product operation applied to each spatial location within the image. Many work~\cite{chen2020pathomic,li2022hfbsurv,wang2021gpdbn} have applied multimodal technique to fusing histology and genomic multimodal data for survival prediction. 
%Wang et al.(2021)~\cite{wang2021gpdbn} propose GPDBN to model complex inter-modal relationships to make full use of the inherent relationships of different modal features for more accurate prognosis prediction of breast cancer.
However, there is no work focusing on the multimodal of the most common stains used in assessment, such as H\&E and IHC,  in cancer multi-class classification.

%design modality-specific attentional factorized bilinear module (MAFB) and cross-modality attentional factorized bilinear module (CAFB) to capture modality-specific relations and leverage high-level fusion for survival prediction.

\section{Method}

\subsection{Two-stage Multimodal bilinear pooling framework}

%ur two-stage framework is shown in Fig~\ref{fig1}. In the first stage, there are bilinear structure of attention-based multiple instance learning pipeline. Each pipeline is trained for extracting bag-level feature within each modality that is beneficial for multi-class classification prediction.

%For multimodal data in cancer pathology, there exists a data heterogeneity gap in combining H\&E and IHC input. The preprocessing of WSI, especially dropping white background of WSI, affects the training performance a lot. In the same pair, preprocessed H\&E image $X_{he} $ may have different size of bag compared to IHC image $X_{ihc} $. Since our training pipeline in first stage extracts bag-level representation $z_n  \in \mathbb{R}^{1\times512}$ for each bag $X$, no matter how the number of instances changed in each bag, the bag-level representation from each modality is always in same size.

%After constructing features from each modality, fusion module is desgined for capturing the robust and informative inter-modality interactions between H\&E and IHC features and fusing new feature representation for the downstream task in the second stage. The second stage have the training pipeline in the same struture with the pipeline in the first stage. We train each stage and each pipeline seperately, making our whole framework training faster than other multimodal model in histopathology field. 

Our framework consists of two stages, as illustrated in Fig~\ref{fig1}. In the first stage, we employ a bilinear structure with attention-based multiple instance learning pipelines. These pipelines are trained to extract bag-level features within each modality, which are beneficial for multi-class classification prediction.
Combining histopathological data from different modalities, such as H\&E and IHC, poses a significant challenge due to the heterogeneity of the data. The preprocessing of WSI, specifically removing the white background, greatly impacts the training performance. Additionally, in the same pair, the preprocessed H\&E image ($X_{he}$) may have a different bag size compared to the IHC image ($X_{ihc}$). However, since our training pipeline in the first stage extracts bag-level representations ($z_n \in \mathbb{R}^{1\times512}$) for each bag $X$, the bag-level representation from each modality remains the same size, regardless of any changes in the number of instances within each bag.

Once the features have been constructed from each modality, we design a fusion module to capture robust and informative inter-modality interactions between H\&E and IHC features. This module constructs the new feature representations as the input of the second stage. The training pipeline in the second stage follows the same structure as the one in the first stage. We train each stage and each pipeline separately, resulting in a faster training process compared to other multimodal models in the histopathology field.

\subsection{Attention-based multiple instance learning training pipeline}

We first describe the whole training process for one branch and subsequently expand upon it to encompass a multi-branch structure.
\subsubsection{The training process for a single branch}
Suppose we have a WSI image $X$, which will denoted as "bag" in the following description. The bag $X$ is split into $K$ patches $X = \{x_{1}, x_{2}, \cdots, x_{K} \} $ , which is denoted as "instance". Each instance will be preprocessed and then fed into ResNet50~\cite{resnet} to get the embedding. Let $R = \{r_{1}, r_{2}, \cdots, r_{K} \} $ be the embedding result from feature extraction for bag $X$ of $K$ instance, $ r_{k} \in \mathbb{R}^{1\times1024}$.  
The first fully-connected layer $W_{fc} \in \mathbb{R}^{512\times1024}$ compresses each embedding $r_{k}$ to $h_k$, a denser feature vector.

\begin{align}
    \boldsymbol{h}_{k} = W_{fc} \boldsymbol{r}_{k}^{T} \in \mathbb{R}^{1\times512}
\end{align}

The denser feature vector $h_k$ is then fed into the multi-class classification network, which consists of attention module and slide-level classfiers. Given $n$ class, the attention network will be split into $n$ parallel $ W_{atten,1}, W_{atten,2}, \\
\cdots, W_{atten,n} \in \mathbb{R}^{1\times256}$. For multi-class classification, the attention module in Attention-Based MIL aggregates the set of embeddings $h_{k}$ into a bag-level embedding $z_{n}$ by Eq.~\ref{equation:zn}, where the attention for the k-th instance is computed by Eq.~\ref{equation:gated-attention-multiclass}.
\begin{equation}
% \sigma_{attention} = \sum_{k = 1}^{K} a_{k,n}h_{k}
\boldsymbol{z}_{n} = \sum_{k = 1}^{K} a_{k,n}\boldsymbol{h}_{k}
\label{equation:zn}
\end{equation}

\begin{equation}
a_{k,n} = \frac{ \exp { \{ W_{atten,n}( \tanh  ( V h_{k}^{T}  ) \odot  sigm ( Uh_{k}^{T} )) \}} }
{\sum_{j = 1}^{K} { \exp { \{ W_{atten,n}( \tanh  ( V h_{j}^{T}  ) \odot  sigm ( U h_{j}^{T} )) \}} } }\qquad
\label{equation:gated-attention-multiclass}
\end{equation}

%\my{
where $a_{k,n}$ is the confidence that $ k^{th}$ instance belongs to $ n^{th}$ class, which denoted as "attention score". $U, V\in \mathbb{R}^{256\times512}$ are learnable attention backbone shared for each class in the attention mechanism, and $\odot $ is the element-wise multiplication for the gated attention mechanism. $z_{n}\in \mathbb{R}^{1\times512}$ is the weighted sum of input $h_k$ for the $ n^{th}$ class. In this way, the input feature  $R = \{r_{1}, r_{2}, \cdots, r_{K}\} $ extracted by ResNet are encoded into a dense feature vector $z_n$ as bag-level representation. We emphasize that the instance number in the bag would not influence the output shape of $z_n$, as each instace-embedding is computed with its attention score to generate the bag-level representation for the $n^{th}$ class.
%which would be discussed and illustated in detail in supplementary material.
%which introduces the possibility for xxxxx,

%where  $a_{k,n}$ is the attention score for $ k^{th}$ instance for the $ n^{th}$ class, $V\in \mathbb{R}^{256\times512}$ , and $U\in \mathbb{R}^{256\times512}$ are learnable attention backbone shared for each class in the attention mechanism,  $\odot $ is the element-wise multiplication for the gated attention mechanism, and the $z_{n} \in \mathbb{R}^{1\times512}$ is the bag-level representation for the $ n^{th}$ class. In this way, different groups of instance feature vectors$ \{r_{1}, r_{2}, \cdots, r_{K} \} $ are learned, and represented by $z_{n}\in \mathbb{R}^{1\times512}$ no matter how the number of instances changed in each bag. Here, $z_{n}$ is also known as the head, as the mechanism is known as multi-head attention.

Then, for the original complete training pipeline, the bag-level representation $z_{n}$ is utilized for predicting the bag-level (also called slide-level) score $s_{n} $. The bag-level score is computed by a group of classifiers $\{W_{c,1},\cdot,W_{c,n} \}$ as Eq\label{equation:CLAM}.(\ref{equation:wc}):
\begin{equation}
s_{n} = W_{c,n} z_{n}^{T}
\label{equation:wc}
\end{equation}
The bag-level score $s_{n} $ will be further fed into $softmax(s_{n})$ for the final prediction.

\subsubsection{Multi-branch multimodal bilinear framework}
The above represents a complete training pipeline for a single branch, as depicted in Fig~\ref{fig1}. For the multimodal processing purpose, we employ two separate pipelines for H\&E and IHC whole slide images in the first stage. Suppose there are M such H\&E and IHC pairs constituting the dataset $  D =\{X_{he,m}, X_{ihc,m},Y_{m}\} ^M_{m=1}$, $Y$ is the label of the H\&E and IHC whole slide image pairs. 
In our paper, we extract the sets of bag representations $\{z_{n}\}^M_{m=1}$ for the dataset $D$ before calculating score $s_{n}$ and $softmax(s_{n})$ in the first stage. 
We construct the sets of $\{z_{he,n}\}^M_{m=1}$ from H\&E-based training pipeline and the sets of $ \{z_{ihc,n}\}^M_{m=1}$ from IHC-based training pipeline, feeding into fusion module to get new feature representations of the whole H\&E and IHC pair as the input of the second stage.

\subsubsection{Loss}

The loss function for each independent branch is:
\begin{equation}
L_{total} = c_1 L_{slide} + c_2 L_{patch}
\end{equation}

where $L_{slide}$ is compared with the ground truth by cross-entropy loss and $L_{patch} $ is the binary smooth SVM loss between the pseudo label generated for top-k patches with the inference of instance-level classifier.

\subsection{Fusion Module}

%For the fusing module,  an average pooing $\sigma $ is employed to reduce the size of the feature space. Then, we apply Kronecker Product to fusion features getting from H\&E-based training pipeline and IHC-based training pipeline to get the new fusion feature $F$.

%~\cite{chen2020pathomic}.

We are motivated to pursue multimodal learning because we believe that the interaction between H\&E and IHC image features will enhance the accuracy of predicting cancer grading. In our previous training pipeline, we created two bag-level feature representations. Now, our goal is to explicitly capture the significant pairwise feature interactions through a fusion module.
%The fusion method is described in the following section 3.3 in how to construct a multimodal representation $F$. 

\subsubsection{Bilinear Average Pooling}

The initial Bilinear pooling mechanism~\cite{kim2016hadamard} was introduced to merge features obtained from two distinct CNNs, both trained on the same image, in order to improve fine-grained visual recognition. The scenario discussed in this paper bears resemblance to the aforementioned situation. As stated earlier, we utilize $X_{he}$ and $X_{ihc}$ as our inputs, representing different staining techniques applied to cells at identical locations, resulting in distinct feature representations of the same sample. Therefore, at this stage, we employ bilinear pooling to fuse these features together.

In contrast to the original Bilinear pooling mechanism~\cite{kim2016hadamard}, which utilizes element-wise product and sum-pooling, our approach employs an average pooling $\sigma$ for bilinear pooling. This is done to reduce the size of the feature space and facilitate the fusion of bilinear features. In the first stage, when dealing with the features $z_{he,n}$ and $z_{ihc,n}$ from the bilinear pipeline, direct convolution fusion is not suitable due to the high dimensionality of these features. By employing average pooling $\sigma$, we are able to compress the dimensions of $z_{he,n}$ and $z_{ihc,n}$ from 512 to 32. Subsequently, these compressed features are fused using the Kronecker product.

\subsubsection{Kronecker product}

The original method proposed by~\cite{kim2016hadamard} utilizes element-wise products to combine features in the Bilinear pooling mechanism. 
In histopathology, previous research~\cite{chen2022pan,chen2020pathomic} has explored various forms of the Kronecker product to fuse histology or genomics features. In other multimodal studies on cancer prognosis prediction, researchers have employed direct feature combination~\cite{huang2019salmon} or score-level fusion~\cite{sahasrabudhe2020deep} to integrate data from different modalities. However, these approaches may not adequately capture the intricate relationships between modalities.

%The original Bilinear pooling mechanism~\cite{kim2016hadamard} applies element-wise product to fuse features. 
%Pervious work in histopathology ~\cite{chen2022pan,chen2020pathomic} has applied the Variations of Kronecker product for the fusion of histology or genomics features. Other previous multimodal studies for cancer prognosis prediction have used direct feature combination\cite{huang2019salmon} or score level fusion\cite{sahasrabudhe2020deep} to integrate data from different modalities, which may not be sufficient to capture the complex inter-modality relations
%Since their scene is to improve performance on fine-grained visual recognition, the element-wise product perfectly demonstrates the process of linear fusion of two features at the same position. 

We utilize the Kronecker product for feature fusion. Consider there are two matrices, $A\in \mathbb{R}^{m\times n}$ and $B\in \mathbb{R}^{l\times p}$. The Kronecker product, denoted as $\otimes$, is an operation that combines these matrices into a block structure, determined by the elements of the original matrices.

$$
	A \otimes B =
	\left[
	\begin{array}{ccc}
	a_{11}B&\cdots& a_{1n}B \\ 
 \vdots&\ddots&\vdots\\
	a_{m1}B&\cdots& a_{mn}B
	\end{array}
	\right]
	$$

Therefore, the joint multimodal tensor constructed from the Kronecker product, as shown in Eq.(\ref{equation:kron}), will capture the important bi-modal interactions that characterize both modalities.

\begin{equation}
F = \sigma (z_{he,n})\otimes\sigma (z_{ihc,n})
\label{equation:kron}
\end{equation}

Upon constructing this joint representation, we utilize the multi-modal representation $F$ as input. We then proceed to learn the representation through the previous training pipeline~\cite{lu2021data}, and subsequently train it using cross-entropy loss for cancer grading tasks. The ultimate significance of pathological fusion lies in its ability to combine diverse patterns and improve prediction accuracy.

\section{Experiments}

\subsection{Datasets}
We use two public breast cancer datasets in this paper. BCI dataset~\cite{bci} presents 4870 registered H\&E and IHC pairs, covering a variety of HER2 expression levels from 0 to 3. IHC4BC dataset~\cite{ihc4bc} contains H\&E and IHC pairs in ER and PR breast cancer assessment, and categories are defined ranges 0 to 3 respectively. The number of each subset is 26135 and 24972.

%Many biomarkers have been explored for their association with treatment response in cancer. For example, the level of estrogen receptor (ER), progesterone receptor (PR), and human epidermal growth factor receptor 2 (HER2) are used to subgroup breast cancer patients and predict their prognosis and response to targeted therapy. 
%The estrogen receptor (ER) and progesterone receptor (PR) are expressed in more than 75% of breast cancers [5, 6]. They are one of the most powerful prognostic factors and predictive markers in hormonal treatment [7–9]. 
%https://www.ncbi.nlm.nih.gov/pmc/articles/PMC9187286/#B9

\subsection{Implementation Details}
We use CLAM~\cite{lu2021data} pre-processing tools to create patches and extract features from each WSI image. Some WSIs will be dropped due to the segment and filtering of CLAM pre-processing mechanism, we take the intersection of H\&E and IHC pre-processed WSIs for further training. The learning rate of Adam optimizer is set to $2 \times 10^{-4}$, the weight decay is set to $1 \times 10^{-5}$, the early-stop stategy is used and the max training epochs is 200. We evaluate our method in two datasets for breast cancer grading tasks with 5-fold Monte Carlo cross-validation. For each cross-validated fold, we randomly split each dataset into 80\%-10\%-10\% subset of training, validation, and testing, stratified by each class. 

\subsection{Results}

%In this paper, we proposed a two-stage multimodal bilinear pooling framework based on Kronecker product to classify multi-stage classifications of HER2, ER and PR in breast cancer. 
In this section, We compared our framework with the following methods frequently used in supervised and weakly supervised learning to test our model’s prediction performance and robustness. Accuracy(ACC) and Area under the ROC Curve (AUC) are selected as evaluation metrics. 

Table~\ref{tab1} illustrated the results of our model compared with other models in BCI dataset.

\vspace{-5pt}
\begin{table}[h]
\caption{Experiment results on BCI dataset.}\label{tab1}
\centering
%\begin{tabular}{l|l|l}
\begin{tabular}{M{4cm}|M{3cm}|M{2cm}|M{2cm}}
\hline
Model & image type& AUC & ACC\\
\hline
InceptionV3 & H\&E &  0.8233 & 0.804\\
Resnet &H\&E & 0.8857 & 0.872\\
Vit~\cite{ayana2023vision} & H\&E & 0.92 & 0.904\\
IHCNet~\cite{shovon2023ihcnet}& IHC&- &0.9345 \\
DenseNet & H\&E & 0.89 & 0.68 \\
HE-HER2Net~\cite{he-her2net}& H\&E& 0.98 & 0.87\\
convoHER2\cite{convoHER2}& H\&E & -& 0.851\\
convoHER2& IHC & -& 0.878\\

CLAM~\cite{lu2021data} & H\&E &0.987 & 0.909\\
CLAM & IHC &0.991&0.917 \\
Two-Stage Multimodal Bilinear Fusion(ours) & H\&E and IHC& \textbf{0.996} & \textbf{0.953}\\
\hline

\end{tabular}

\end{table}
%densenet he-her2net \cite{he-her2net}

% vit: resnet inceptionv3
\vspace{-10pt}

As shown in Table~\ref{tab1}, our two-stage bilinear fusion framework outperformed all other existing models, achieving an impressive AUC of 0.996 and ACC of 0.953. Our model achieved a remarkable AUC score close to 1, indicating its strong classification ability among multiple classes. Hence, our proposed model exhibited exceptional performance in the multi-class classification task.

%, a precision of 0.88, a recall of 0.86.

We also conducted experiments on the IHC4BC-ER and -PR datasets to test our motivation and methods. Our findings demonstrate that IHC-based machine model performs better than H\&E-based model. Specifically, our two-stage bilinear fusion framework achieves a higher ACC compared to both H\&E-based and IHC-based models in the IHC4BC dataset.

\begin{table}[t]
\caption{Experiment results on IHC4BC dataset.}\label{tab2}
\centering
\begin{tabular}{M{5cm}|l|l|l|l|l}
\hline
\multirow{2}{*}{Model}& \multirow{2}{*}{image type} &\multicolumn{2}{|c|}{IHC4BC-ER} &\multicolumn{2}{|c}{IHC4BC-PR} \\
& &  AUC & ACC & AUC & ACC \\
\hline
CLAM& H\&E &  0.9543	&0.8421&  0.9089&	0.7734	 \\
CLAM&IHC &  0.9796	&0.8941& 0.9609	&0.847	\\
Two-Stage Multimodal Bilinear Fusion(ours) & H\&E and IHC&0.981&0.905&  0.9621 &	0.8482 \\

\hline
\end{tabular}
\end{table}

%\textbf{the reason why we apply three CLAM model in our framework is that we want to exclude the impact of the performance differences caused by training model} CLAM model in three training pipeline can be changed into other CNN-based model. 

%\textbf{our framework is not end-to-end training. Each training pipeline only contains 0.79 million trainable parameters. Since we are two-stage bilinear framework, our framework only need to train 0.79 millon parameters each time, which is much smaller than other end-to-end deep bilinear network\cite{wang2021gpdbn}.}

\subsection{Discussion}

In this section, we analyze the wrong prediction of each model and derive two key insights:
\subsubsection{IHC medical images have inherent characteristics in color and morphology}
The H\&E WSIs depicted in Fig. \ref{fig2} (a) exhibit incorrect predictions, contrasting with the accurately predicted paired IHC WSIs. The third column illustrates the IHC staining level related to the grade of the WSI by color deconvolution. The fourth column presents a 3D scatter plot of optical density RGB values. These observations underscore the color characteristics in IHC images enabling the IHC-based model to learn representation well.

%As shown in Fig\ref{fig2} (a), these H\&E images, which have ground truth from 1 to 3, are predicted wrong while inferencing. However, their paired IHC images have specific IHC staining color characteristics. 
%The label of the first row of H\&E WSI is 1, the second row is 2, and the third one is 3. The third column in (a) is the deconv of IHC images, and the fourth column is the 3D scatter plot of optical density RGB values. A wide range of color changes and staining intensities in blue and brown regions can be observed in different level IHC WSIs.  

\subsubsection{H\&E-based model learn wrong high attention area or can't learn the right area well}
\begin{figure}[t]
\includegraphics[width=\textwidth]{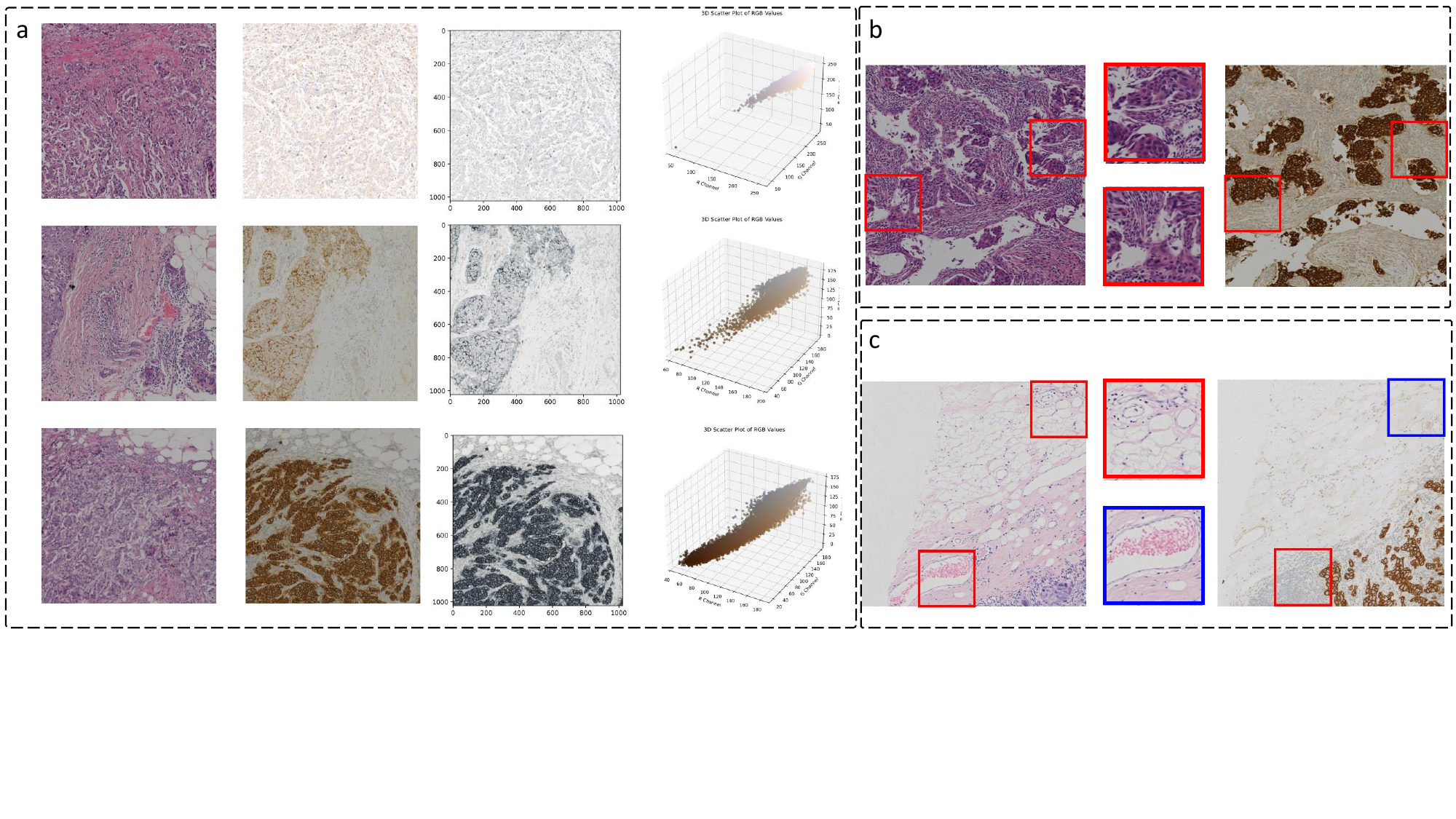}
\caption{H\&E-based machine learning the wrong prediction. However, for their paired IHC images, there are huge differences in color and can be identified correctly in the IHC-training pipeline.} \label{fig2}
\end{figure}

H\&E-based models allocate top-k attention correctly but struggle to extract meaningful representations from H\&E images. The CLAM model, using clustering and attention-based MIL mechanism, sometimes allocates the highest top-k attention correctly but still makes wrong predictions. Additionally, in Fig. \ref{fig2} (c), the trained model allocates the highest top-k attention to erroneous areas, as evidenced by the zoom-in patch in the blue box. These areas contain fewer staining cells with brown hues in the paired IHC image, leading to inaccurate predictions.
%H\&E-based models demonstrate a propensity to allocate top-k attention to the appropriate areas, but their predictive accuracy is compromised due to the model's inability to extract meaningful representations from H\&E images. As shown in Fig. \ref{fig2} (b), in certain test cases, the trained CLAM model, which employs top-k clustering and attention-based MIL mechanism, allocates the highest top-k attention to the correct area, containing cells with intense staining in paired IHC images, yet yields wrong predictions. In additional, in Fig. \ref{fig2} (c), the trained model allocates highest top-k attention to erroneous areas, as evidenced by the zoom-in patch in the blue box. These areas contain fewer staining cells with brown hues in the paired IHC image, consequently leading to inaccurate predictions.

\section{Conclusion}

We present a novel two-stage multimodal bilinear pooling fusion framework, designed for integrating both H\&E as well as IHC whole slide images into cancer grading tasks. Our experiments on the publicly available BCI dataset demonstrate that our framework achieves state-of-the-art performance. The outcomes of our study underscore the efficacy of employing multimodal features in histopathological tasks, resulting in enhanced performance metrics. Moreover, our experiments reveal that aligning IHC datasets with H\&E at the pixel level is unnecessary. 
%These results emphasize the importance of integrating IHC analysis into existing computational pathology workflows based on H\&E, thereby advocating for a more robust predictive framework in cancer diagnostics.

%We propose two-stage multimodal bilinear pooling fusion framework for both H\&E and IHC whole slide image analysis in cancer grading tasks. Our experiments on the public dataset BCI produce the state-of-the-art performance. Our experiments prove that IHC matters in histopathological tasks in improving performance, and demonstate that IHC dataset do not need to aligh to H\&E in pixel level. These findings highlight the importance of integrating IHC analysis into current H\&E-based computational pathology workflows, advocating for a more robust predictive framework in cancer diagnostics.

\bibliographystyle{splncs04}
\bibliography{mybibliography}

\end{document}